\pdfoutput=1

\documentclass[11pt]{article}

\usepackage[]{acl}

\usepackage{times}
\usepackage{latexsym}

\usepackage[T1]{fontenc}

\usepackage[utf8]{inputenc}

\usepackage{microtype}

\usepackage{inconsolata}

\usepackage{times}
\usepackage{latexsym}
\usepackage{times}
\usepackage{latexsym}
\usepackage{graphicx}
\usepackage{lipsum}

\usepackage{float}
\usepackage{algorithm}
\usepackage{algorithmic}
\usepackage{placeins}
\usepackage{subfigure}
\usepackage{booktabs}
\usepackage{cleveref}
\usepackage{multirow}
\usepackage{xcolor, soul}
\definecolor{green}{HTML}{C2D8C8}
\definecolor{purp}{HTML}{D2C2D8}
\definecolor{red}{HTML}{F42C14}
\definecolor{dark-red}{HTML}{DE3163}
\definecolor{dark-blue}{HTML}{268EEA}
\sethlcolor{red}

\title{Analyzing the Evaluation of Cross-Lingual Knowledge Transfer \\ in Multilingual Language Models}

\author{Sara Rajaee \and Christof Monz \\
  Language Technology Lab, University of Amsterdam \\
  \texttt{\{s.rajaee, c.monz\}@uva.nl} 
  }

\begin{document}
\maketitle
\begin{abstract}
Recent advances in training multilingual language models on large datasets seem to have shown promising results in knowledge transfer across languages and achieve high performance on downstream tasks.
However, we question to what extent the current evaluation benchmarks and setups accurately measure zero-shot cross-lingual knowledge transfer.
In this work, we challenge the assumption that high zero-shot performance on target tasks reflects high cross-lingual ability by introducing more challenging setups involving instances with multiple languages.
Through extensive experiments and analysis, we show that the observed high performance of multilingual models can be largely attributed to factors not requiring the transfer of actual linguistic knowledge, such as task- and surface-level knowledge.
More specifically, we observe what has been transferred across languages is mostly data artifacts and biases, especially for low-resource languages. 
Our findings highlight the overlooked drawbacks of existing cross-lingual test data and evaluation setups, calling for a more nuanced understanding of the cross-lingual capabilities of multilingual models.\footnote{The source code is available at: \url{https://github.com/Sara-Rajaee/crosslingual-evaluation}}

\end{abstract}

\section{Introduction}
Massively Multilingual Transformers (MMTs) exhibit remarkable abilities in comprehending texts in multiple languages \cite{devlin-etal-2019-bert, conneau-etal-2020-unsupervised, chi-etal-2022-xlm, lin-etal-2022-shot}.
Their performance on extensive multilingual benchmarks \cite{ hu2020xtreme, ruder-etal-2021-xtreme} including natural language inference \cite{conneau-etal-2018-xnli}, paraphrase identification \cite{yang-etal-2019-paws}, question answering \cite{artetxe-etal-2020-cross}, and commonsense reasoning \cite{ponti-etal-2020-xcopa}, clearly indicates their usefulness for downstream tasks across different languages. 

Since collecting labeled data for multiple languages is expensive, fine-tuning an MMT on only one language (usually English) and applying it to other languages is a common practice to expand multilingual models' applications to more languages. The success of employing this approach is primarily attributed to the cross-lingual ability of MMTs and knowledge transfer across languages \cite{wang2019cross,keung-etal-2020-dont,fujinuma-etal-2022-match, ebrahimi-etal-2022-americasnli}. 
However, in this work, we question this attribution and take a closer look at the cross-lingual evaluation pipelines used to assess MMTs' performance. Our study aims to broaden our horizons about the existing performance-based evaluation setups of MMTs and highlight their significant shortcomings.

To this end, we first provide insights into the definition of the cross-lingual ability in language models and the essential criteria for its evaluation.
We challenge the assumption of \textit{high} cross-lingual ability, i.e., requiring actual linguistic knowledge, in MMTs using three downstream tasks: Natural Language Inference (NLI), Paraphrase Identification (PI), and Question Answering (QA).
Our experimental results demonstrate that multilingual models struggle with transferring linguistic knowledge across languages when the inputs involve multiple languages, such as in the NLI task with
a premise in Arabic and a hypothesis in Spanish.
Employing this setup, we show that, unlike previous assumptions, MMTs are not able to effectively connect the underlying semantics between languages in a zero-shot manner.
It is worth mentioning that the evaluation of MMTs using multiple languages in the input offers both theoretical advantages and reflects real-world scenarios in NLP systems.

We extend our study to investigate if the failures come from the lack of \textit{across}-language fine-tuning data.
We find that even by fine-tuning MMTs on \textit{across}-language data that involves two languages in an instance, they still can not successfully transfer knowledge between languages in a zero-shot setting.

Looking for the reasons behind the ineffectiveness of MMTs on the \textit{across} setup, 
we examine the impact of individual samples on the cross-lingual performance and identify a specific subset where MMTs struggle. 
Our findings demonstrate that models achieve exaggerated high performance by strongly relying on spurious features and data artifacts. 
We notice that cross-lingual transfer primarily involves learned biases and shallow knowledge rather than linguistic knowledge. 
Notably, this phenomenon disproportionately affects low-resource languages, exacerbating the challenges faced by MMTs in achieving true cross-lingual competence.

As part of our methodology, we design control tasks \cite{hewitt-liang-2019-designing} in which, during fine-tuning, the words within the instances have been randomly shuffled, and then, the model is evaluated on the original, i.e., not shuffled, test data.
Surprisingly, our experiments show that although these new tasks do not provide the model with meaningful linguistic knowledge related to the target task, there is only a slight drop in their cross-lingual performance in both single and two-language evaluation settings.
These results demonstrate that MMTs' understanding tends to be more reliant on surface-level patterns rather than linguistic comprehension.

Our experiments show from several angles that current MMTs' cross-lingual evaluation setups do not give us a clear and faithful picture of their cross-lingual ability. 
Our findings question the extent of high crosslinguality in language models and prompt us to pay more attention to the interpretation of knowledge transfer and cross-lingual ability in multilingual models solely based on their performance on downstream tasks.

\section{Related Work}

\paragraph{Massively Multilingual Transformers.}

Multilingual models have achieved success in understanding multiple languages without requiring language-specific supervision, thanks to only requiring unlabelled training data in multiple languages.
Prominent examples include Multilingual BERT (mBERT) and XLM-R, which have been pre-trained on a diverse set of languages using the masked language modeling objective \cite{devlin-etal-2019-bert,conneau-etal-2020-unsupervised}. 
Other models, such as XLM with Translation Language Modeling (TLM), XLM-E with multilingual replaced token detection (MRTD) and translation replaced token detection (TRTD), and ALM with code-switched inputs, have employed similar strategies with different objectives \cite{NEURIPS2019_c04c19c2,chi-etal-2022-xlm,Yang_Ma_Zhang_Wu_Li_Zhou_2020}.

\paragraph{Analyzing MMT.}

The rise of multilingual models has sparked significant interest to understand their linguistic capabilities across different languages. 
\citet{chi-etal-2020-finding}  have discovered subspaces within mBERT representations that can capture syntactic tree distances across different languages. 
Aligned with their finding, numerous studies have further explored the potential of multilingual models to capture both syntactic \cite{papadimitriou-etal-2021-deep, xu-etal-2022-cross, mueller-etal-2022-causal,ravishankar-etal-2021-attention} and semantic knowledge \cite{foroutan-etal-2022-discovering, vulic-etal-2020-probing} for a wide range of languages.

Several studies have delved into understanding the factors influencing the cross-lingual ability of multilingual models. 
\citet{pires-etal-2019-multilingual} explored the effectiveness of multilingual BERT in transferring knowledge across languages using NER and POS tagging tasks, noting the impact of language similarity on performance. 
\citet{chai-etal-2022-cross} examined cross-linguality from a language structure perspective, emphasizing the significance of the composition property in facilitating cross-lingual transfer.
\citet{muller-etal-2021-first} analyzed representation similarities and discovered a strong connection between hidden cross-lingual similarity and the model's performance on downstream tasks. 
Building upon this finding, \citet{deshpande-etal-2022-bert} identified a correlation between token embedding alignment and zero-shot transfer across diverse tasks.

In the realm of multilingual models, both the design of new models and the examination of existing ones rely heavily on evaluating their performance in zero-shot cross-lingual scenarios. 
However, there remains the question of how accurately we can interpret their performance and its implications for the models' cross-lingual abilities. 
To shed light on this issue, our work focuses on investigating the faithfulness of the prevailing methods used to evaluate zero-shot cross-lingual performance in the literature. 
By doing so, we aim to provide deeper insights into the interpretation of model performance and its relationship to the cross-lingual capabilities of multilingual models.

\section{Cross-lingual Evaluation} 

The ability of a multilingual model to effectively generalize across languages on downstream tasks is the key factor in determining its cross-linguality \cite{pires-etal-2019-multilingual, wu-dredze-2019-beto,artetxe-etal-2020-cross}. 
More specifically, a multilingual model is considered cross-lingual if it can successfully perform tasks in languages not seen during task fine-tuning.

However, relying solely on the ultimate performance of multilingual models on a specific target task has significant drawbacks, as it can potentially create misconceptions about their true cross-lingual abilities. 
This evaluation approach lacks clarity in distinguishing between the extent of cross-lingual knowledge transfer and surface-level and non-linguistic features.
It is possible for a multilingual model to achieve high performance on a task without possessing a deep semantic understanding of a language, instead mostly relying on language-independent and shallow knowledge, as previously reported for monolingual language models \cite{bhargava-etal-2021-generalization, stacey-etal-2020-avoiding,gururangan-etal-2018-annotation,mccoy-etal-2019-right}.

It is worth mentioning that fully disentangling linguistic and shallow (task-specific) knowledge is almost impossible for most NLP tasks. 
However, trying to separate these types of knowledge as much as possible gives us a clearer and more accurate perspective regarding the linguistic knowledge captured by models across languages. 

In the following parts, we employ three different target tasks, namely multilingual Natural Language Inference, Paraphrase Identification, and Question Answering, to study the cross-lingual ability of multilingual models. 
We utilize multilingual BERT (\citealp[mBERT]{devlin-etal-2019-bert}) and the base version of XLM-r \cite{conneau-etal-2020-unsupervised}
for our experiments as they are among the most widely used multilingual models.
We also employ I{\small{NFO}}XLM \cite{chi-etal-2021-infoxlm} trained on a cross-lingual objective and parallel data to investigate the role of explicit pre-trained cross-lingual objective on the cross-lingual ability.\footnote{The experimental setups of fine-tuning are provided in the Appendix.}

\subsection{Natural Language Inference}
\begin{table*}[ht!]
    \centering   
    \setlength{\tabcolsep}{6.5pt}
    \scalebox{0.8}{
    \begin{tabular}{r | c c c c c c c c c c c c c c c |c}

    \toprule
              & en & de & fr & ru & es & zh & vi & ar & tr & bg & el & ur & hi & th & sw & avg \\
    \midrule          
    \multicolumn{17}{c}{\textbf{mBERT}}
    \\
    \cmidrule{2-16}
    \it within & 81.5  & 
            70.6   & 
            73.5   & 
            68.6   & 
            68.2   &
            68.6   & 
            69.9   & 
            64.2   & 
            62.0   & 
            68.7   & 
            67.5   & 
            58.7   & 
            60.5   & 
            52.3   & 
            50.3   &
            65.7\\
    \it across &  61.3 & 
                57.7   &
                59.5   &
                57.2   &
                59.2   &
                55.2   &
                56.0   &
                54.3   &
                51.1   &
                55.9   &
                54.3   &
                50.7   &
                52.3   &
                47.2   &
                45.6   &
                54.5 \\

    \midrule
    \multicolumn{17}{c}{\textbf{XLM-r}}
    \\
    \cmidrule{2-16}
    \it within & 84.9  & 
            76.3   & 
            78.3   & 
            75.7   & 
            79.2   &
            73.5   & 
            74.8   & 
            71.5   & 
            73.0   & 
            78.6   & 
            75.4   & 
            65.5   & 
            69.3   & 
            71.8   & 
            65.2   &
            74.2\\
    
    \it across &  71.9 & 
                67.1   &
                68.8   &
                68.0   &
                69.2   &
                64.6   &
                65.1   &
                62.8   &
                62.8   &
                68.3   &
                66.2   &
                60.0   &
                63.6   &
                64.2   &
                53.7   &
                64.8\\

    \midrule
    \multicolumn{17}{c}{\textbf{I{\small{NFO}}{XLM}}}
    \\
    \cmidrule{2-16}
    \it within & 85.8  & 
            78.2   & 
            79.2   & 
            76.9   & 
            80.0   &
            75.5   & 
            75.9   & 
            73.2   & 
            74.4   & 
            78.4   & 
            77.0   & 
            66.0   & 
            71.0   & 
            73.0   & 
            65.9   &
            75.4\\
    
    \it across &  77.1 & 
                72.0   &
                72.9   &
                71.9   &
                73.2   &
                70.0   &
                69.9   &
                69.0   &
                69.0   &
                72.5   &
                71.1   &
                64.8   &
                68.8   &
                69.8   &
                61.8   &
                70.3\\
    \bottomrule

    \end{tabular}}

    \caption{The accuracy scores of mBERT, XLM-r, and I{\small{NFO}}{XLM} for two evaluation settings on the NLI task: the \textit{within} language setting, where both the hypothesis and premise are in the same language, and the \textit{across} language evaluation, which involves two different languages. In the \textit{across} evaluation, numbers represent the average performance when either the premise or the hypothesis (but not both) is in the given language. As the numbers show, MMTs have considerably lower performance in the \textit{across} setting.}
    \label{tab:xnli-performance}
\end{table*}
\begin{table}[ht!]
    \centering   
    \setlength{\tabcolsep}{4.5pt}
    \scalebox{0.8}{
    \begin{tabular}{r | c c c c c c c |c }

    \toprule
              & en & de & fr & es & zh & ko & ja & avg \\
    \midrule 
    \multicolumn{9}{c}{\textbf{mBERT}}
    \\
    \cmidrule{2-8}
    \textit{within}
    & 93.5
    & 84.6
    & 86.6
    & 86.7
    & 77.0
    & 72.8
    & 73.6
    & 81.4 \\
    \textit{across}
    & 75.1
    & 72.1
    & 72.5
    & 72.6
    & 64.8 
    & 64.1 
    & 63.7
    & 69.2 \\
    \midrule 
    \multicolumn{9}{c}{\textbf{XLM-r}}
    \\
    \cmidrule{2-8}
    \textit{within}
    & 94.4
    & 87.8
    & 89.5
    & 89.1
    & 81.9
    & 76.3
    & 77.3
    & 85.7\\
    \textit{across}
    & 76.4
    & 72.1
    & 72.2
    & 72.2
    & 63.6
    & 65.2
    & 62.9
    & 69.2\\
    \midrule 
    \multicolumn{9}{c}{\textbf{I\small{NFO}\large{XLM}}}
    \\
    \cmidrule{2-8}
    \textit{within}
    & 94.0
    & 88.4
    & 90.0
    & 90.2
    & 83.0
    & 78.7
    & 78.9
    & 86.2\\
    \textit{across}
    & 84.6
    & 79.9
    & 80.3
    & 80.4
    & 75.2
    & 72.4
    & 73.3
    & 78.0\\
\bottomrule
\end{tabular}}

    \caption{The accuracy of fine-tuned models on PAWS-X evaluated under the \textit{within} and \textit{across} language settings. For mBERT and XLM-r, there is, on average, a 17\% drop in performance when the sentences are provided in two different languages. While the trend is similar for I{\small{NFO}}XLM, the drop is less significant in this model, which can be attributed to its cross-lingual pretraining objective.}
    \label{tab:paws-performance}
\end{table}

\begin{table*}[ht!]
    \centering   
    \setlength{\tabcolsep}{4.5pt}
    \scalebox{0.8}{
    \begin{tabular}{r | c c c c c c c c c c c| c}
    \toprule
            & en & de & ru & es & zh & vi & ar & tr & el & hi & th & avg\\
    \midrule          
    \multicolumn{13}{c}{\textbf{mBERT}}
    \\
    \cmidrule{2-12}
    \it within & 84.5  & 
           72.7    & 
            71.4   & 
            75.5   &
            58.2   & 
            69.2   & 
            61.2   & 
            55.1   & 
            62.4   & 
            58.0   &
            40.0   &
            64.4 \\
    \it across &  57.0  & 
                50.6    &
                50.2    &
                52.7    &
                42.6    &
                46.3    &
                42.2    &
                36.9    &
                42.4    &
                38.7    &
                26.1    &
                44.2   \\

        \midrule          
    \multicolumn{13}{c}{\textbf{XLM-r}}
    \\
    \cmidrule{2-12}
    \it within & 84.2  & 
            75.2   & 
            74.5   & 
            77.1   & 
            63.7   &
            74.3   & 
            66.3   & 
            68.1   & 
            73.8   & 
            68.3   &
            66.5   &
            72.0   \\
    \it across &  58.1 & 
                44.1   &
                42.9   &
                43.4   &
                28.1   &
                33.9   &
                25.8   &
                31.3   &
                34.7   &
                32.3   &
                29.6   &
                36.8   \\
    \midrule          
    \multicolumn{13}{c}{\textbf{I\small{NFO}\large{XLM}}}
    \\
    \cmidrule{2-12}
    \it within & 85.1  & 
            76.0   & 
            75.0   & 
            77.8   & 
            66.4   &
            75.6   & 
            70.3   & 
            69.9   & 
            74.8   & 
            71.6   &
            69.9   &
            73.8   \\
    \it across &  73.8 & 
                66.6   &
                66.7   &
                68.1   &
                61.4   &
                64.3   &
                61.0   &
                61.4   &
                64.2   &
                61.9   &
                60.1   &
                64.5   \\
    \bottomrule

 \end{tabular}}

    \caption{Zero-shot F1 scores of fine-tuned models for the QA task using the \textit{within} and \textit{across} language evaluation approaches. All the models struggle with the \textit{across} setup, especially for mBERT and XLM-r, where we observe more than $50\%$ drop in their performance, challenging the extent of their cross-linguality.}
    \label{tab:qa-performance}
\end{table*}

Natural language inference serves as a prominent task in the field of NLP for assessing the comprehension capabilities of language models \cite{bowman-etal-2015-large, condoravdi-etal-2003-entailment}. 
This task requires the prediction of the relationship between a given premise and hypothesis, where the model determines whether the premise entails the hypothesis, contradicts it, or remains undetermined \cite[MNLI]{williams-etal-2018-broad}.

To study the cross-lingual ability of MMTs in the NLI task, we employ the multilingual NLI dataset \cite[XNLI]{conneau-etal-2018-xnli}, where the training set has 397k samples in English (adopted from the MNLI training set), and the test sets include 5k instances in fifteen different languages manually translated from the English data \cite{pmlr-v119-hu20b}.

We investigate the cross-linguality of MMTs in the context of NLI using two distinct evaluation settings. 
In the first setting, which we refer to as the \textit{within language} setup, we fine-tune the model on English training data and assess its performance on NLI tasks across other languages, following previous studies \cite{artetxe-schwenk-2019-massively,xlm, wu-dredze-2019-beto,qi-etal-2022-enhancing,chai-etal-2022-cross,huang-etal-2021-improving-zero}. 
In the second setting, the \textit{across language} evaluation, we assess cross-linguality by providing premise and hypothesis pairs in two different languages. 
To the best of our knowledge, none of the previous research has evaluated the cross-linguality of MMTs by employing instances involving multiple languages in downstream tasks.
Since the XNLI test data is a fully parallel dataset, we can easily combine premises and hypotheses from different languages.
We assert that this evaluation approach provides a more precise and reliable assessment of the models' cross-linguality. 
It asks the model to comprehend the underlying meaning of the input in two languages simultaneously, minimizing potential spurious correlations between examples and labels. 
It is worth mentioning that both evaluation approaches are conducted in a zero-shot manner.

The results are presented in Table~\ref{tab:xnli-performance}. 
For a comprehensive breakdown of performance for each language pair, please refer to the Appendix.
Aligned with previous research \cite{pmlr-v119-hu20b}, the results of the \textit{within} language setting show the ability of MMTs to effectively generalize knowledge learned during fine-tuning from English to other languages when the premise and hypothesis are in the same language. 
Nevertheless, the extent of their success varies across languages, with lower accuracy observed for low-resource languages like Swahili compared to high-resource ones.

The results of the \textit{across} language experiment, where the premise and hypothesis are in two different languages, show a comparatively lower level of cross-linguality in MMTs compared to the \textit{within} setup.
It raises concerns about the true extent of cross-lingual ability in language models. 
Even for high-resource languages within the same language family (e.g., English and German), the average performance declines by approximately 17\% for mBERT and XLM-r. This drop is even more pronounced for low-resource languages such as Swahili, see Figures \ref{fig:xnli-mbert-performance-detail}--\ref{fig:xnli-info-performance-detail}.
As we expected, I{\small{NFO}}XLM exhibits less of a performance drop in the \textit{across} setup attributed to its cross-lingual pre-training objective. 
However, since English has been used as a pivot language in the pre-training data, the cross-lingual objective has mostly helped the performance of pairs including English, and the trend of the other pairs is similar to mBERT and XLM-r, see Figure \ref{fig:xnli-info-performance-detail}.

\subsection{Paraphrase Identification}
The Paraphrase Identification task evaluates a model's understanding of the semantic similarity between two sentences \cite{wang-etal-2018-glue}.  
Paraphrase Adversaries from Word Scrambling (PAWS) is a challenging dataset for this task, where both sentences in each example have high word overlap \cite{zhang-etal-2019-paws}. 
PAWS-X is a multilingual benchmark and extends this dataset to six languages beyond English using professionally translated validation and test sets \cite{yang-etal-2019-paws}. 

We employ a similar evaluation setup as described for NLI to assess the cross-lingual capability of multilingual models on the semantic similarity task. 
Since not all instances are translated into all six other languages in the dataset, we only consider parallel sentences for the evaluation, resulting in the exclusion of a small number of examples (less than $0.5\%$ on average) from the test sets.

The performance of the fine-tuned models on PAWS-X is presented in Table~\ref{tab:paws-performance}, and the detailed results can be found in Figures~\ref{fig:paws-mbert-performance-detail}--\ref{fig:paws-info-performance-detail}. 
In the \textit{within} language setup, where the sentences in every instance are from the same language, the models demonstrate the successful knowledge transfer across languages, as indicated by their relatively high performance compared to English. 
However, in the \textit{across} language setting, which tests the models' cross-lingual ability in a more challenging scenario, similar to the findings in XNLI, there is a significant drop in performance for the models. 
These results show that comprehending information from two different languages and making semantic-based decisions pose challenges for multilingual models, particularly for non-Latin script languages, where their performance is noticeably affected.

\subsection{Question Answering}
The question answering (QA) task challenges the reading comprehension ability in language models in which the model is asked to find the answer span to a question within the given context \cite{choi-etal-2018-quac,rajpurkar-etal-2016-squad}. 
While NLI and PI are similarity-based and classification tasks, QA offers an alternative perspective to evaluate the cross-lingual capacity of MMTs in a different type of target task.
For this task, we employ XQuAD \cite{artetxe-etal-2020-cross}, a multilingual question-answering benchmark comprising 240 paragraphs and 1,190 question-answer pairs from the development set of SQuAD v1.1 \cite{rajpurkar-etal-2016-squad}. 
It includes professional translations into ten languages,
making it a fully parallel dataset.

The results of the cross-lingual evaluation of the QA models are presented in Table~\ref{tab:qa-performance} for the \textit{within} and \textit{across} language setups.\footnote{The language-pair results can be found in the appendix and see Table~\ref{tab:qa-performance-em} for EM performance.}
We report the exact match (EM) and F1 scores, which are commonly used metrics to evaluate question-answering systems. The reported results for the \textit{across} setting are the average over cases that the context or question (not both of them) are in the listed language.  

For the QA task, we observe a similar trend to the NLI and PI tasks.
The numerical results indicate that multilingual models excel in generalizing across unseen languages when the context and question are in the same language, i.e., the \textit{within} setting. 
However, when the context is in a different language than the question, the empirical results show a substantial decline in performance. 
This suggests that multilingual models face challenges in retrieving knowledge from several languages at the same time and bridging information across different language representation spaces.

Another interesting observation is that in the \textit{across} language scenario, the models' performance is lower compared to both languages in the \textit{within} language setup. 
Nonetheless, there is an exception when the question is in English in the \textit{across} language setting. 
Regardless of whether the context language is low or high-resource language, the performance remains close to the \textit{within} setup of the context language in mBERT and XLM-r (Figures~\ref{fig:qa-mbert-performance-detail} and \ref{fig:qa-xlme-performance-detail}).
We speculate that this pattern may be linked to the utilization of Wikipedia articles in the pre-training data for both models, which were used in constructing the SQuAD dataset as well.
However, we observe an opposite pattern for the I{\small{NFO}}XLM model in that the model relies more on the context than the question that can be attributed to the English-centric pretraining data.

\subsection{Discussion}
\begin{table}[ht!]
    \centering   
    \setlength{\tabcolsep}{6.5pt}
    \scalebox{0.85}{
    \begin{tabular}{r | c c c c}

    \toprule
              & en-de & ar-en & tr-el & es-de  \\
    \midrule          
    \textit{$l1-l2$} & 67.8 $_{\small{64.1}}$  & 
            75.9 $_{\small{64.1}}$  & 
            63.4 $_{\small{52.9}}$  & 
            76.4 $_{\small{61.8}}$  
            \\
    \textit{$l2-l1$} &  58.7 $_{\small{71.3}}$ & 
                39.2  $_{\small{56.5}}$ &
                41.7  $_{\small{50.8}}$ &
                43.1  $_{\small{65.3}}$ 
                 \\
    \textit{$*-l2$} &  63.0 $_{\small{59.4}}$& 
                59.6   $_{\small{55.0}}$&
                61.7   $_{\small{52.7}}$&
                63.2   $_{\small{58.7}}$
                 \\
    \textit{$l1-*$} &  57.6 $_{\small{58.7}}$& 
                41.6   $_{\small{59.4}}$&
                55.2   $_{\small{55.0}}$&
                43.8   $_{\small{58.7}}$
                \\
    \textit{$*-*$} &  52.8 $_{\small{55.3}}$& 
                54.9   $_{\small{55.3}}$&
                56.5   $_{\small{55.3}}$&
                56.4   $_{\small{55.3}}$
                \\

\bottomrule
    
    \end{tabular}}

    \caption{mBERT's performance on the XNLI test set. The columns show the fine-tuning language pairs, and the rows show the evaluation pairs in which $l1$ and $l2$ represent the premise and hypothesis's languages, respectively, as given in the corresponding columns. The rows that include $*$ show the average performance over all languages. The smaller numbers present the baseline performance (fine-tuning on en-en) for the corresponding evaluation pairs.}
    \label{tab:xnli-mix-mbert}
\end{table}

The previous experimental results demonstrate that MMTs face difficulties in understanding and connecting multiple languages simultaneously. 
Although the cross-lingual pre-training objective of I{\small{NFO}}XLM has enhanced the cross-lingual ability of this model compared to mBERT and XLM-r, the performance gap between \textit{within} and \textit{across} setting is still considerable.
A question raised here is whether the low performance comes from the lack of \textit{across} language style fine-tuning data or a deeper incapability of cross-lingual knowledge transfer in MMTs. 
To answer this question, we fine-tune mBERT using \textit{across} language setups on multiple language pairs on the NLI task. 
Table~\ref{tab:xnli-mix-mbert} breaks down the performance for the zero-shot and semi-zero-shot setups.\footnote{The results of other language pairs can be found in the appendix.} As can be observed, fine-tuning using \textit{across} language style data does not help the model to generalize to other languages. 
Interestingly, the model cannot even achieve better performance on the same language pair when the premise and hypothesis languages have been swapped.
Considering the results, we conclude that the lower performance on the \textit{across} evaluation setup can not be attributed to the distinct fine-tuning and evaluation setups.

\begin{table*}[ht!]
    \centering   
    \setlength{\tabcolsep}{6.5pt}
    \scalebox{0.8}{
    \begin{tabular}{r | c c c c c c c c | c c c c c c c |c}

    \toprule
              & en & de & fr & ru & es & zh & vi & ar & tr & bg & el & ur & hi & th & sw & avg \\
    \midrule          
    \multicolumn{17}{c}{\textbf{Entailment}}
    \\
    \it within & 76.9  & 
            61.2   & 
            66.8   & 
            60.4   & 
            66.6   &
            54.8   & 
            61.7   & 
            56.6   & 
            62.7   & 
            63.0   & 
            62.0   & 
            46.3   & 
            51.9   & 
            60.1   & 
            68.1   &
            61.3\\
    \it across &  32.2 & 
                28.1   &
                31.3   &
                28.4   &
                30.5   &
                21.0   &
                24.0   &
                22.2   &
                16.2   &
                26.0   &
                22.1   &
                15.2   &
                19.5   &
                10.3   &
                ~8.3   &
                22.3 \\

    \multicolumn{17}{c}{\textbf{NotEntailment}}
    \\
    \it within & 83.9  & 
            75.3   & 
            76.9   & 
            72.8   & 
            77.7   &
            75.0   & 
            73.5   & 
            68.0   & 
            61.6   & 
            71.6   & 
            70.2   & 
            64.9   & 
            64.8   & 
            48.5   & 
            41.4   &
            67.3\\
    
    \it across &  75.9 & 
                72.5   &
                73.7   &
                71.6   &
                73.5   &
                72.2   &
                71.6   &
                70.3   &
                68.5   &
                70.7   &
                70.4   &
                68.4   &
                68.6   &
                65.7   &
                64.2   &
                70.5\\
    \bottomrule

    \end{tabular}}

    \caption{mBERT's accuracy scores on the XNLI test set separated based on the labels. While both classes have an almost equal contribution to the performance in the \textit{within} setup, the performance on the \textit{entailment} class significantly drops in the \textit{across} setup.}
    \label{tab:xnli-perlabel-mbert}
\end{table*}

\begin{table}[ht!]
    \centering   
    \setlength{\tabcolsep}{4.5pt}
    \scalebox{0.8}{
    \begin{tabular}{r | c c c c c c c |c }

    \toprule
              & en & de & fr & es & zh & ko & ja & avg \\
    \midrule 
    \multicolumn{9}{c}{\textbf{Paraphrase}}
    \\

    \textit{within}
    & 94.9
    & 89.0
    & 90.1
    & 88.8
    & 76.5
    & 55.7
    & 67.4
    & 81.4 \\
    \textit{across}
    & 57.6
    & 54.7
    & 54.9
    & 53.3
    & 35.5 
    & 33.0 
    & 31.2
    & 45.7 \\
    \midrule 
    \multicolumn{9}{c}{\textbf{NonParaphrase}}
    \\

    \textit{within}
    & 92.4
    & 81.0
    & 83.8
    & 85.0
    & 77.5
    & 86.6
    & 78.6
    & 82.9\\
    \textit{across}
    & 89.3
    & 86.1
    & 86.8
    & 88.3
    & 88.6
    & 89.3
    & 89.8
    & 88.3\\
    
\bottomrule
\end{tabular}}

    \caption{The performance of mBERT for the \textit{within} and \textit{across} setting per label on the PI task. Most of the performance drop of the \textit{across} setting occurs for the \textit{paraphrase} class.}
    \label{tab:paws-mbert-performance}
\end{table}

\section{Breakdown Analysis}

In this section, we delve into the reasons behind the lack of success exhibited by multilingual models in the \textit{across} language setup. 
Our analysis begins by exploring the individual contributions of each class to the overall tasks' performance. 
By doing so, we aim to uncover the specific instances that present difficulties for multilingual models and subsequently lead to a decline in their performance.

\paragraph{NLI.}
In Table~\ref{tab:xnli-perlabel-mbert}, we present the numerical results for the NLI task per label.\footnote{Please refer to Table~\ref{tab:xnli-perlabel-xlmr} and \ref{tab:xnli-perlabel-info} for the XLM-r and I{\small{NFO}}XLM results.} 
Following previous work \cite{yaghoobzadeh-etal-2021-increasing,sanh2021learning}, and for simplicity, we combine the \textit{neutral} and \textit{not-entailment} classes, considering them as the \textit{not-entailment} class. 
Surprisingly, the contribution of the \textit{entailment} and \textit{not-entailment} classes to the overall performance in the \textit{across} language setup is not equal, and this trend is consistent across all models. 
Notably, the drop in performance is primarily affected by the entailment class and is more pronounced for low-resource languages.

In relation to this behavior, we suspect that the observed patterns can be attributed to dataset artifacts, particularly the word overlap bias in the training set of XNLI \cite{mccoy-etal-2019-right}, and this bias is transferred to other languages. 
The word overlap bias refers to the tendency of NLI models trained on the MNLI dataset to favor the \textit{entailment} label when there is a high word overlap between the premise and hypothesis. 
Moreover, previous studies have shown a strong correlation between low word overlap and the \emph{not-entailment} label, which is referred to as reverse (word overlap) bias \cite{rajaee-etal-2022-looking}. Therefore, since in the \textit{across} setting, the word overlap is minimized, the model is biased toward predicting the \textit{not-entailment} label.
Another possible explanation is that the multilingual models prioritize language similarity over semantic meaning, leading to a considerable drop in performance.

In the multilingual context, our findings show that, in the case of the NLI task, the transfer across languages primarily involves dataset artifacts and biases rather than linguistic knowledge.
Especially in low-resource scenarios, where sufficient pre-training data is lacking, the model heavily relies on these shortcuts.

\paragraph{PI.}
The performance of mBERT on the PAWS-X test sets, broken down by labels, is presented in Table~\ref{tab:paws-mbert-performance} (see the other models' results in Table~\ref{tab:paws-xlmr-performance} and \ref{tab:paws-info-performance}). 
Despite the adversarial construction of the PAWS dataset to prevent word overlap bias, with all instances having high word overlap between sentences, we observe a similar drop in performance in the cross-language setting, primarily originating from the \textit{paraphrase} class. 
This suggests the possible presence of biases that have been transferred across languages rather than linguistic knowledge transfer.

To understand this behavior, we consider two potential reasons for the drop in \textit{paraphrase} class performance.
Firstly, the overlap bias may arise from the pre-training procedure of multilingual models. 
Additionally, there might be fine-grained biases related to bigram or trigram overlap in the PAWS training data. 
Investigating these intriguing patterns and biases in the dataset is an avenue for future research.

\begin{figure*}
    \centering
    \scalebox{0.95}{
    \includegraphics[height = 6cm, width=14.5cm]{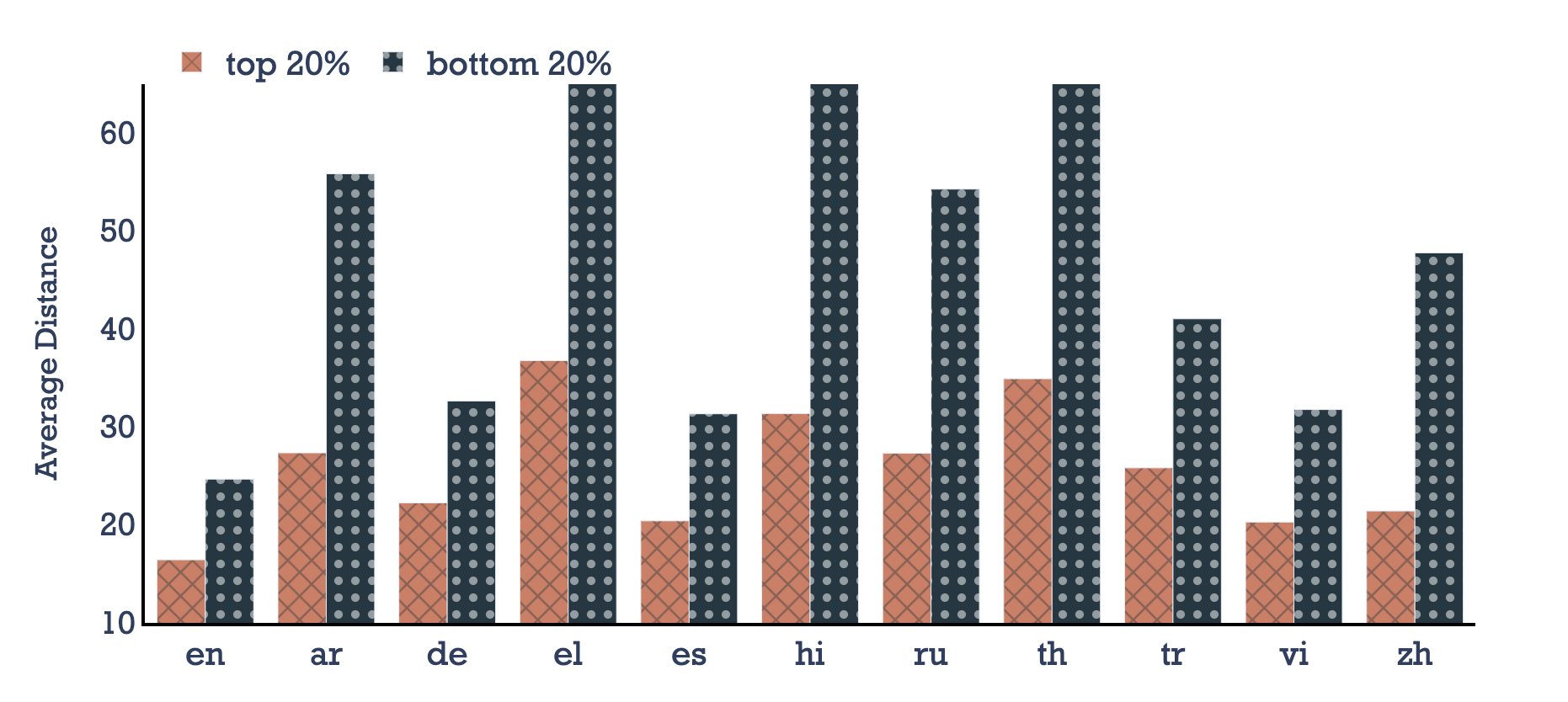}}
    \caption{The average distance of the questions' words occurred in the context to the center of the answer span in the top $20\%$ easiest and hardest instances for the mBERT fine-tuned on SQuAD evaluated on every language. As the average distance increases, the model's performance drops.}
    \label{fig:qa-mbert-distance}
\end{figure*}
\paragraph{QA.}
For the question-answering task, which differs from a classification task, we adopt a distinct approach to uncover the obstacles impeding knowledge transfer across languages in the \textit{across} setup.

Based on the nature of our \textit{across} language setup that minimizes the overlap between the context and question and the results of the NLI analysis, we suspect that a similar bias exists for the QA task, which is easily transferable across languages. 
Previous studies have reported different biases in the SQuAD dataset, including answer-position, word overlap between the question and context, and type-matching \cite{ko-etal-2020-look, sugawara-etal-2018-makes, weissenborn-etal-2017-making}.

To capture the degree of overlap between the context and question, we introduce a new measurement technique. 
Instead of simple word count, we compute the average distance of each question word's occurrence in the context to the center of the answer span.
Here, the difference between the shared word position index in the context and the center of the answer span is considered as the distance. 
If there is no word overlap between the question and context, we assign the distance as the maximum length of context (which is a hyperparameter).
We contend that the model's reliance on high word overlap cannot be simply regarded as a shortcut, and we advocate for the use of the distance metric as a more accurate measure of the possible bias evaluation.

To investigate a possible spurious correlation between the question and context overlap and the answer span, we calculate the average distance for the top and bottom $20\%$ of samples, representing instances where the model achieves the highest and lowest F1 scores, respectively, across all individual languages.

The findings are presented in Figure~\ref{fig:qa-mbert-distance}.\footnote{Please refer to Figures~\ref{fig:qa-xlmr-distance} and \ref{fig:qa-info-distance} for more results.}
It is evident that the average distance for the most challenging instances is twice that of the easiest ones, indicating a significant correlation between the concentration of shared words around the answer span and the model's performance. 
Furthermore, reliance on the concentration of shared words around the answer as a shortcut is consistently observed across different languages, as indicated by their corresponding performances.

\begin{table*}[ht!]
    \centering   
    \setlength{\tabcolsep}{3.9pt}
    \scalebox{0.8}{
    \begin{tabular}{r | c c c c c c c c c c c c c c c c c|c | c}

    \toprule
              & en & de & fr & ru & es & zh & vi & ar & tr & bg & el & ur & hi & th & sw & ko & ja & \textbf{avg} & \textbf{original}\\
    \midrule          
    \multicolumn{20}{c}{\textbf{XNLI}}
    \\
    \cmidrule{2-20}
    \it within & 77.9  & 
            68.7   & 
            71.3   & 
            66.2   & 
            71.0   &
            66.3   & 
            67.0   & 
            62.4   & 
            61.3   & 
            65.9   & 
            64.1   & 
            56.0   & 
            59.0   & 
            51.8   & 
            48.8   &
            ~-~    &
            ~-~    &
            63.8   &
            65.7 \\
    \it across &  57.8 & 
                55.1   &
                56.4   &
                55.0   &
                56.6   &
                52.5   &
                53.4   &
                52.4   &
                49.8   &
                53.4   &
                52.0   &
                49.1   &
                50.6   &
                46.9   &
                44.8   &
                ~-~    & 
                ~-~    &
                52.4   &
                54.5 \\

    \cmidrule{2-20}
    \multicolumn{20}{c}{\textbf{PI}}
    \\
    \cmidrule{2-20}
    \it within & 57.1  & 
            55.5   & 
            55.2   & 
            ~-~    & 
            56.1   &
            54.9   & 
            ~-~    & 
            ~-~    & 
            ~-~    & 
            ~-~    & 
            ~-~    & 
            ~-~    & 
            ~-~    & 
            ~-~    & 
            ~-~    & 
            54.0   & 
            54.8   &

            55.4   &
            81.4   \\
    \it across &  56.8 & 
                56.1   &
                56.0   &
                ~-~    &
                56.3   &
                55.8   &
                ~-~    &
                ~-~    &
                ~-~    &
                ~-~    &
                ~-~    &
                ~-~    &
                ~-~    &
                ~-~    &
                ~-~    &
                55.6   &
                56.1   &
                56.1   &
                69.2   \\

    \cmidrule{2-20}
    \multicolumn{20}{c}{\textbf{QA}}
    \\
    \cmidrule{2-20}
    \it within & 82.5  & 
            68.6   & 
            ~-~    & 
            70.0   & 
            72.9   &
            58.0   & 
            67.6   & 
            56.8   & 
            55.1   & 
            ~-~    & 
            60.0   & 
            ~-~    & 
            56.0   & 
            43.3   & 
            ~-~    &
            ~-~    &
            ~-~    &
            62.8   &
            64.4   \\

    \it across &  54.0 & 
                46.4   &
                ~-~    &
                46.5   &
                48.8   &
                38.4   &
                41.6   &
                37.7   &
                34.9   &
                ~-~    &
                37.3   &
                ~-~    &
                35.5   &
                25.7   &
                ~-~    &
                ~-~    &
                ~-~    &
                40.6   &
                44.2   \\
    \bottomrule

    \end{tabular}}

    \caption{The performance of mBERT fine-tuned on the control tasks. We report accuracy for the NLI and PI tasks and the F1 score for QA. Although the fine-tuning data does not provide the model with any meaningful and task-related knowledge, the drop in performance is negligible for the NLI and QA tasks. The original column refers to the results in Tables~\ref{tab:xnli-performance}--\ref{tab:qa-performance}.}
    \label{tab:shuffle-performance-mBERT}
\end{table*}

\paragraph{Summary.}

Our analysis across different tasks highlights the predominant influence of dataset artifacts and reliance on shortcuts rather than robust cross-lingual knowledge transfer.
The observed disability in performing tasks involving multiple languages shows that the models prioritize shallow knowledge over linguistic understanding.

\section{Control Tasks}

The idea of control tasks, proposed by \citet{hewitt-liang-2019-designing}, is aimed at enabling a meaningful and faithful interpretation of the linguistic knowledge encoded in language models' representations during probing procedures. 
In the interpretability area, these tasks serve as baselines to measure the model's language understanding capabilities and ensure that the probe's high performance is not attributed to the linguistic knowledge learned by the probe itself and coming from the encoded linguistic knowledge in the representations.

In this section, we borrow the idea of control tasks in probing and employ them to assess the cross-lingual abilities of MMTs. 
These tasks are indeed (partially) random sequences, allowing us to evaluate the models' performance when it does not receive linguistic cues.
This unconventional approach provides valuable insights into the extent of the meaningful numerical performance of multilingual models on the current multilingual benchmarks.

To this aim, we randomly shuffle the inputs and fine-tune the models on these shuffled instances.
More specifically, for the NLI and PI tasks, we shuffle the words within every sentence, and for the QA task, we shuffle the question words and keep the original context. 

It is evident that the designed tasks are only marginally related to human language comprehension.
However, surprisingly, the performance of the models on these tasks, as shown in Table~\ref{tab:shuffle-performance-mBERT}, indicates only a marginal decrease for both within and cross-language settings.

In the case of the NLI task, we observe a slight $4\%$ drop in accuracy on the test sets when the model is fine-tuned on a nonsensical task. 
Similarly, the QA model exhibits a drop of approximately $3\%$. 
These findings suggest that the current test sets may not provide sufficient quality for effectively evaluating the cross-lingual capabilities of multilingual models. 
Out of all the tasks, the PI task stands out with its random performance, indicating that the fine-tuned model with shuffled data lacks the necessary knowledge to complete the task successfully. 
This aligns with our expectations from a high-quality dataset.

\section{Future Directions}
In this work, we have demonstrated that current performance-based methods for evaluating cross-lingual abilities are insufficient.
The main concern now is how to assess the cross-lingual capabilities of multilingual models.

Firstly, we should be aware of the shortcomings of existing multilingual benchmarks commonly used for evaluating cross-lingual knowledge transfer.
As we described in our study, current tasks and datasets are prone to artifacts leading to exaggerated high performance. 
In addition to the random baseline, which is defined based on the number of labels for classification tasks, we suggest having a secondary baseline for every task.
The secondary baseline could be considered the performance of a simple neural network on the task or the performance on the control tasks. 
We leave the exploration of this avenue to future work. 

Another alternative could be our suggested setup, the \textit{across} language approach, which involves multiple languages. 
The proposed evaluation setup is not only valuable for theoretical analysis but also reflects real-world scenarios where NLP systems need to integrate knowledge from different language sources.

\section{Conclusion}
In this paper, we take a fresh perspective on the cross-lingual ability of multilingual models. 
Through comprehensive experiments, we explored their capacity to simultaneously leverage knowledge from multiple languages. 
Our findings show that multilingual models struggle to establish connections between knowledge spaces across languages, resulting in subpar performance on cross-language task setups, i.e., when there are multiple languages in the input. 
Our results show that the previously reported high performance in the zero-shot setting predominantly stems from the transfer of shallow, language-independent knowledge. 
Surprisingly, we observed that dataset artifacts, rather than intrinsic linguistic features, are predominantly transferred across languages.
This challenges the notion of relying solely on multilingual models' performance for assessing their true cross-lingual capabilities.
To assess the quality of existing multilingual benchmarks, we conducted fine-tuning experiments on control tasks with nonsensical input. 
Surprisingly, even when fine-tuned on meaningless tasks, the models demonstrated exceptionally high performance, prompting concerns regarding the quality of current multilingual datasets. 
In light of these insights, in the future, we plan to explore novel task and data-independent approaches to gain a more accurate understanding of multilingual models' true cross-lingual abilities. 
\section*{Limitations}

In our experiments, we primarily examined three widely used MMTs, namely mBERT, XLM-r, and I{\small{NFO}}XLM. However, there is room for further expansion by incorporating a broader range of MMTs with diverse objectives and architectures to evaluate their cross-lingual ability.
Additionally, while we assessed the cross-linguality of MMTs on three downstream tasks, there is potential for exploring additional target tasks and datasets. 

\section{Acknowledgements}
We want to thank the anonymous reviewers for their valuable comments and suggestions, which helped us in improving the paper. This research is funded in part by the Netherlands Organization for Scientific Research (NWO) under project number VI.C.192.080.

\bibliography{anthology,custom}

\appendix
\section{Experimental Setups}
We have used the HuggingFace library for fine-tuning multilingual models. For the NLI task, we have fine-tuned the models for 3 epochs with a batch size of $32$, maximum length of $128$, and learning rate of $2e-5$, using the last [CLS] token representation.
For the PI task, we have considered the same hyperparameters but a batch size of $16$. 
For the QA task, we have used the following setups: a maximum length of $384$, batch size of $16$, and learning rate of $2e-5$. All the reported numerical results are the average over three different random seeds. 
All models have been fine-tuned using one NVIDIA A6000 GPU.

\label{sec:appendix}
\begin{figure*}
    \centering
    \scalebox{0.9}{
    \includegraphics[height = 12cm, width=12.5cm]{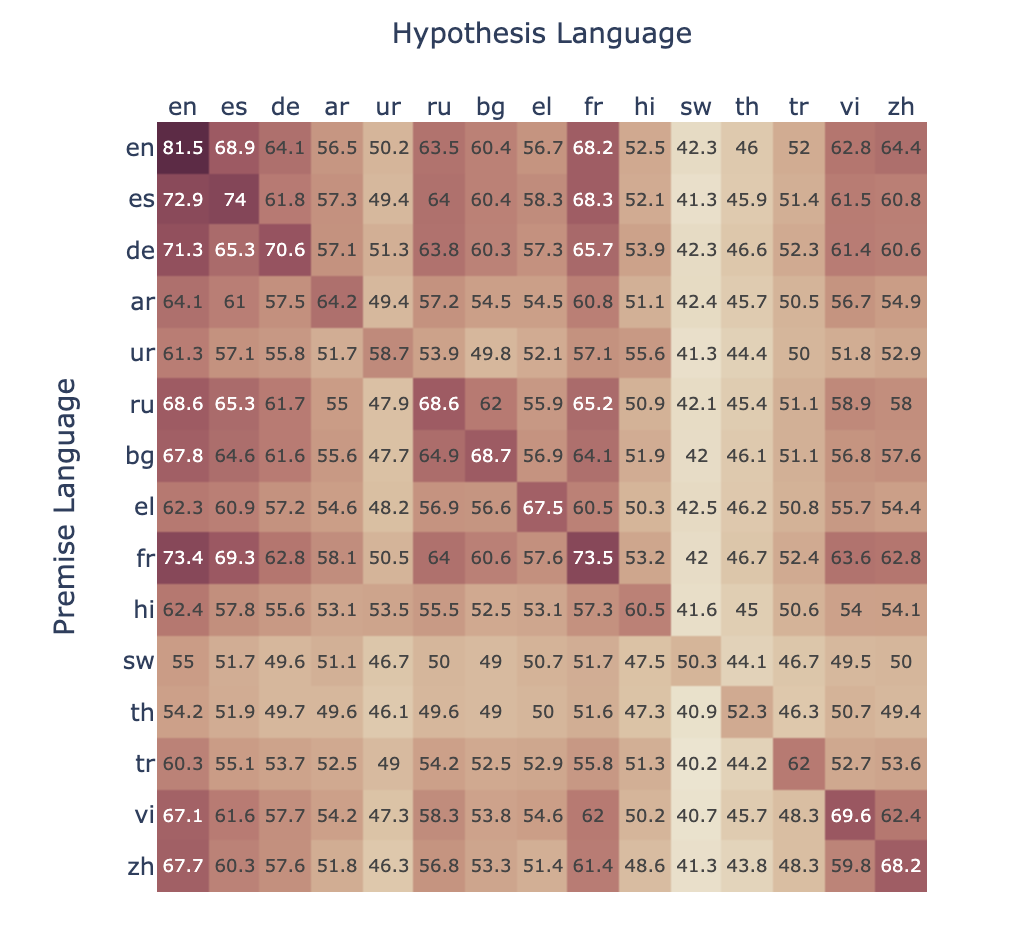}}
    \caption{Language-pairs accuracy scores for mBERT  on the multilingual NLI task.}
    \label{fig:xnli-mbert-performance-detail}
\end{figure*}

\begin{figure*}
    \centering
    \scalebox{0.9}{
    \includegraphics[height = 12cm, width=13cm]{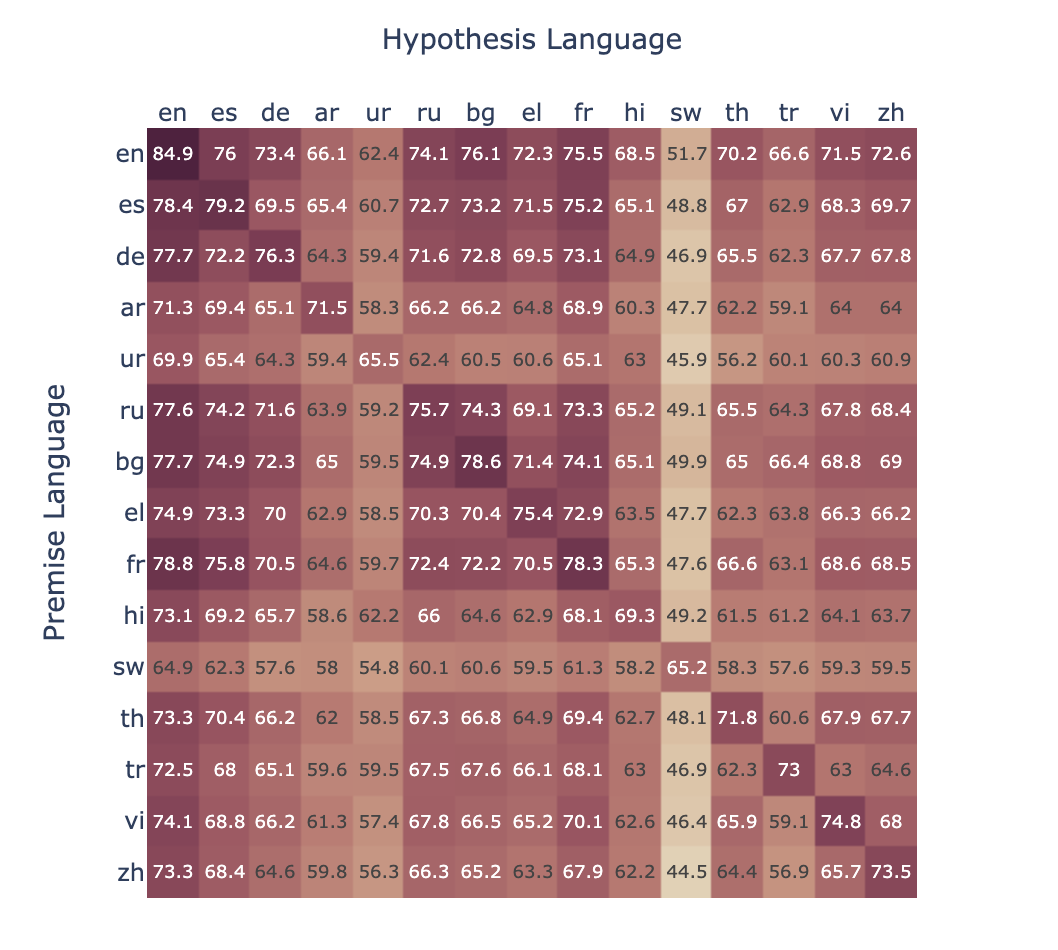}}
    \caption{Language-pairs accuracy scores for  XLM-r  on the multilingual NLI task.}
    \label{fig:xnli-xlmr-performance-detail}
\end{figure*}

\begin{figure*}
    \centering
    \scalebox{0.9}{
    \includegraphics[height = 12cm, width=13cm]{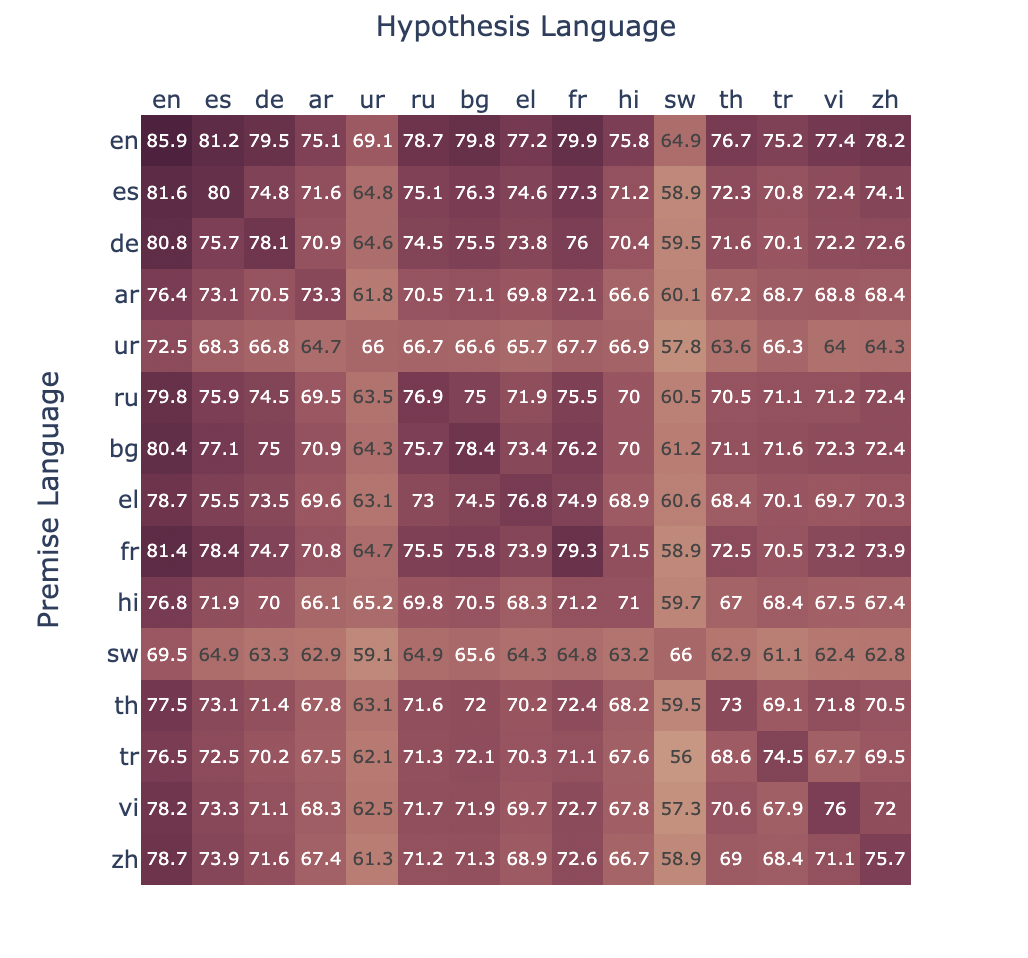}}
    \caption{Language-pairs accuracy scores for  I{\small{NFO}}XLM  on the multilingual NLI task.}
    \label{fig:xnli-info-performance-detail}
\end{figure*}

\begin{figure}
    \centering
    \scalebox{1}{
    \includegraphics[height = 6cm, width=6.5cm]{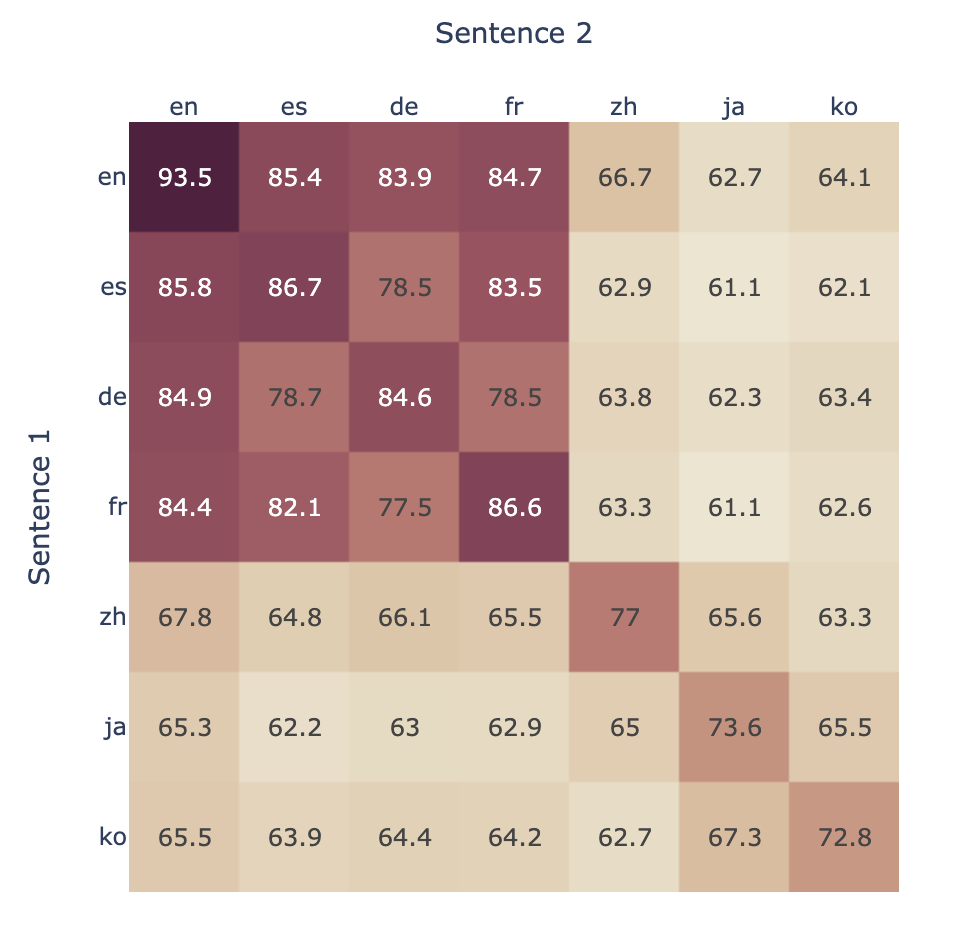}}
    \caption{Language-pairs accuracy scores for  mBERT  on the multilingual PI task.}
    \label{fig:paws-mbert-performance-detail}
\end{figure}
\begin{figure}
    \centering
    \scalebox{1}{
    \includegraphics[height = 6cm, width=6.5cm]{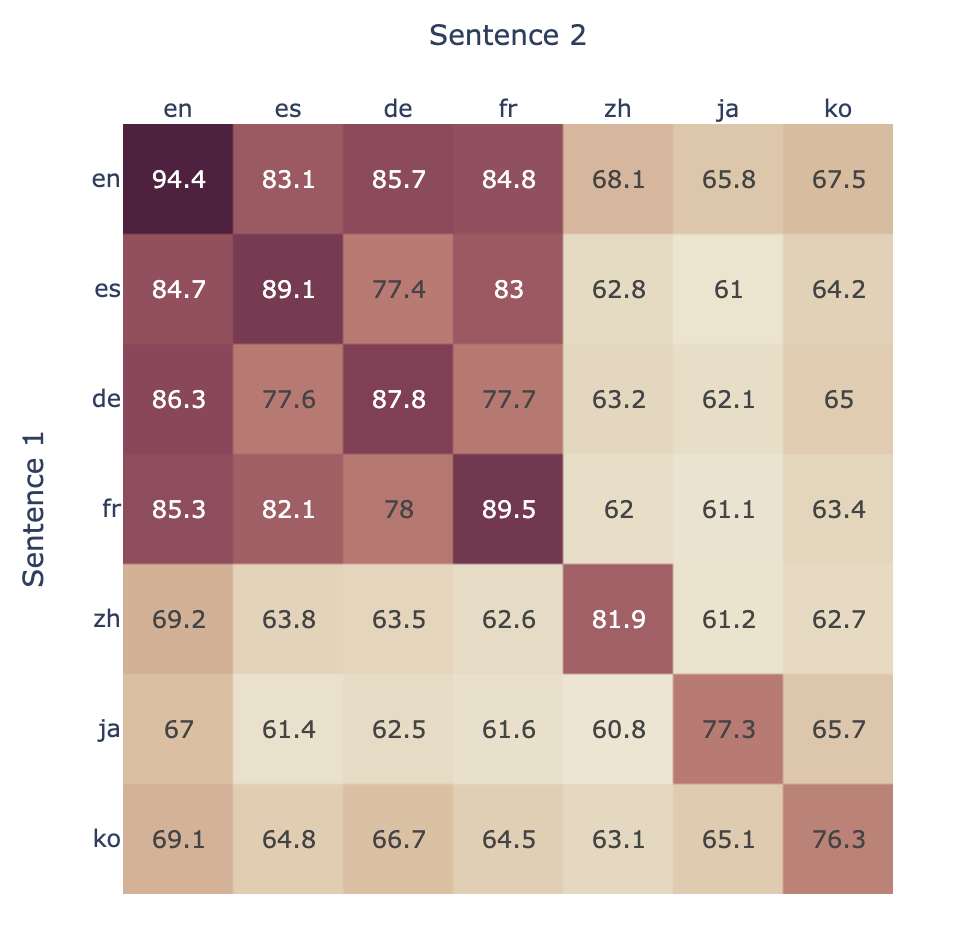}}
    \caption{Language-pairs accuracy scores for  XLM-r  on the multilingual PI task.}
    \label{fig:paws-xlmr-performance-detail}
\end{figure}
\begin{figure}
    \centering
    \scalebox{1}{
    \includegraphics[height = 6cm, width=6.5cm]{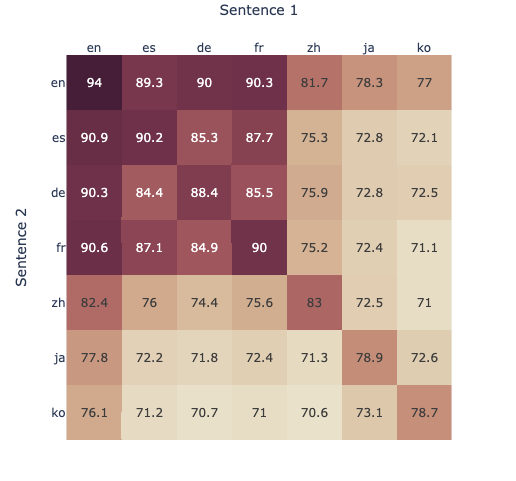}}
    \caption{Language-pairs accuracy scores for I{\small{NFO}}XLM on the multilingual PI task.}
    \label{fig:paws-info-performance-detail}
\end{figure}

\begin{figure*}
    \centering
    \scalebox{0.75}{
    \includegraphics[height = 12cm, width=12.5cm]{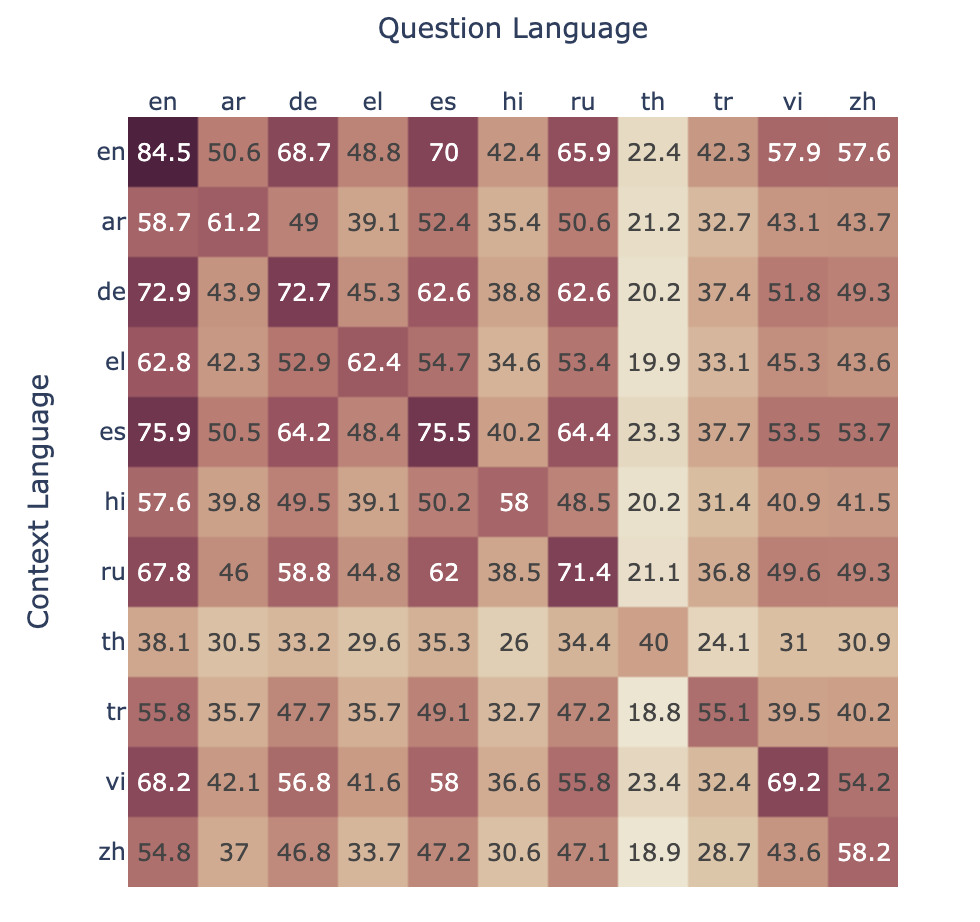}}
    \caption{Language-pairs F1 score for mBERT on the multilingual QA task.}
    \label{fig:qa-mbert-performance-detail}
\end{figure*}

\begin{figure*}
    \centering
    \scalebox{0.75}{
    \includegraphics[height = 12cm, width=12.5cm]{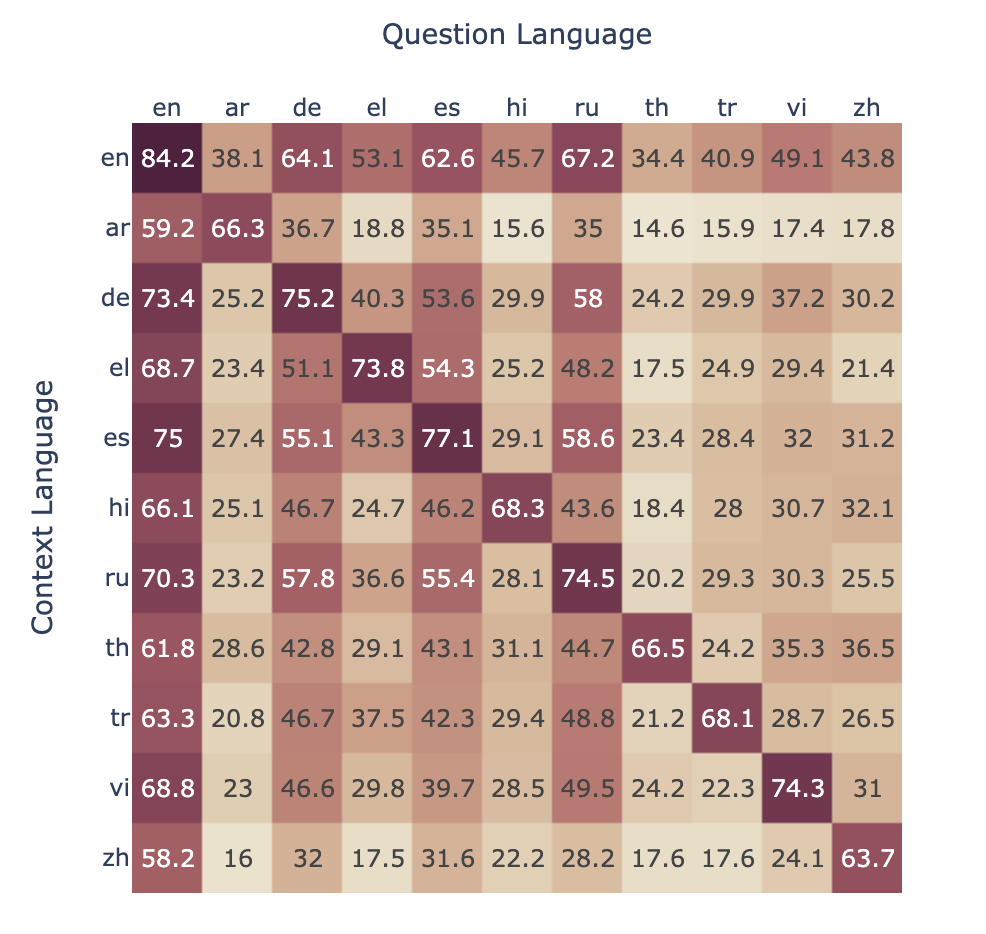}}
    \caption{Language-pairs F1 score for XLM-r on the multilingual QA task.}
    \label{fig:qa-xlme-performance-detail}
\end{figure*}

\begin{figure*}
    \centering
    \scalebox{0.75}{
    \includegraphics[height = 12cm, width=12.5cm]{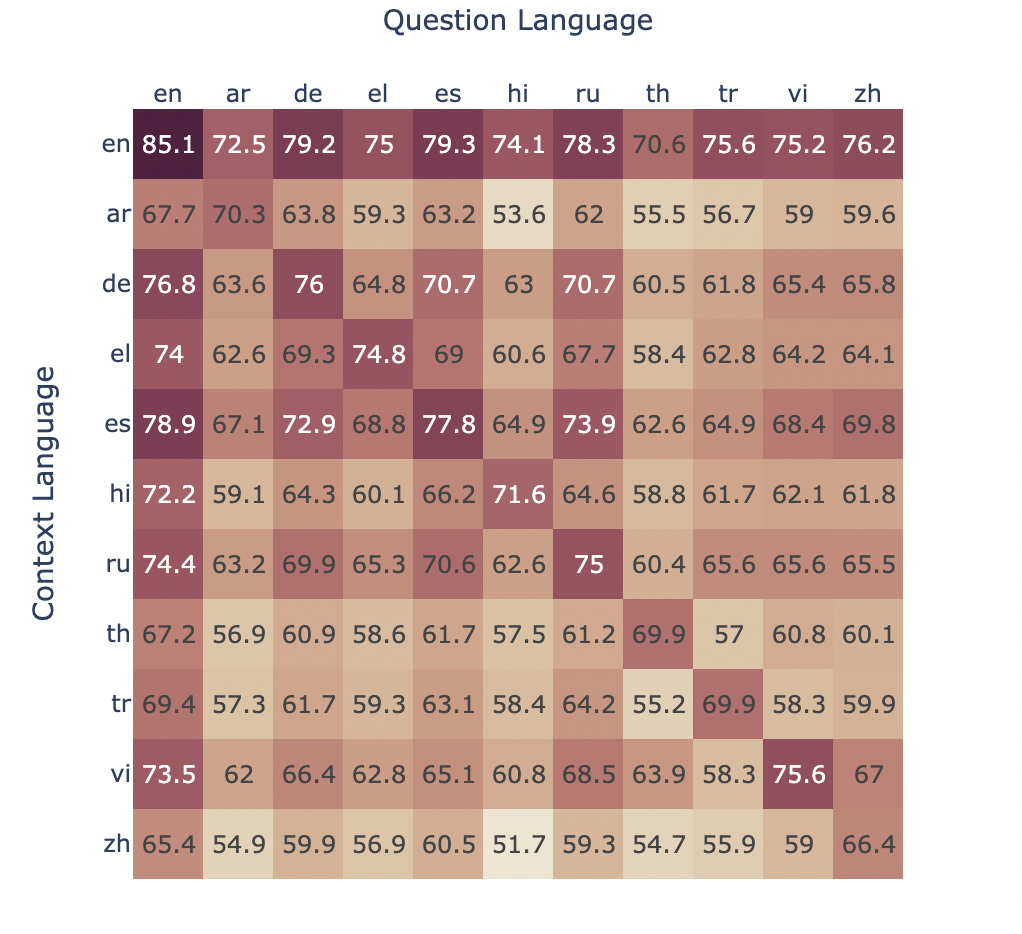}}
    \caption{Language-pairs F1 score for I{\small{NFO}}XLM  on the multilingual QA task.}
    \label{fig:qa-info-performance-detail}
\end{figure*}

\begin{table*}[ht!]
    \centering   
    \setlength{\tabcolsep}{4.5pt}
    \scalebox{0.8}{
    \begin{tabular}{r | c c c c c c c c c c c| c}
    \toprule
            & en & de & ru & es & zh & vi & ar & tr & el & hi & th & avg\\
    \midrule          
    \multicolumn{13}{c}{\textbf{mBERT}}
    \\
    \cmidrule{2-12}
    \it within &  73.0  & 
            56.6 & 
            54.5 & 
            57.0 &
            48.1 & 
            50.1 & 
            44.6 & 
            40.1 & 
            45.2 & 
            43.9 &
            31.3 &
            49.5\\
    \it across &  43.0 & 
                36.5 &
                36.0 &
                37.1 &
                30.0 &
                31.2 &
                28.9 &
                24.8 &
                28.9 &
                26.2 &
                17.3 &
                30.9\\

        \midrule          
    \multicolumn{13}{c}{\textbf{XLM-r}}
    \\
    \cmidrule{2-12}
    \it within & 73.4& 
            59.9& 
            58.0& 
            59.4& 
            53.8&
            54.7& 
            50.4& 
            52.1& 
            56.4& 
            52.0&
            54.3&
            56.8\\
    \it across &  44.0& 
                30.7&
                29.4&
                29.2&
                17.2&
                20.4&
                14.8&
                19.1&
                22.1&
                20.1&
                17.9&
                28.1\\
    \midrule 
                \multicolumn{13}{c}{\textbf{I{\small{NFO}}XLM}}
    \\
    \cmidrule{2-12}
    \it within & 74.1& 
            60.4& 
            58.9& 
            59.4& 
            57.5&
            55.9& 
            53.4& 
            53.8& 
            56.5& 
            52.0&
            54.7&
            58.5\\
    \it across &  60.3& 
                51.6&
                51.4&
                51.8&
                48.7&
                47.7&
                45.4&
                46.6&
                47.9&
                47.0&
                46.4&
                49.5\\
    \bottomrule

 \end{tabular}}

    \caption{Zero-shot EM scores of fine-tuned models for teh QA task using the \textit{within} and \textit{across} language evaluation approaches. There is a more than $50\%$ drop in performance under the \textit{across} evaluation method challenging the extent of cross-linguality of mBERT and XLM-r.}
    \label{tab:qa-performance-em}
\end{table*}

\begin{table}[ht!]
    \centering   
    \setlength{\tabcolsep}{6.5pt}
    \scalebox{0.75}{
    \begin{tabular}{l | c c c c c }

    \toprule
              & \textit{$l1-l2$} & \textit{$l2-l1$}& \textit{$*-l2$} &\textit{$l1-*$} & \textit{$*-*$} \\
    \midrule          
    \textit{en-de} & 67.8 $_{\small{64.1}}$  & 
            58.7 $_{\small{71.3}}$  & 
            63.0 $_{\small{59.4}}$  & 
            57.6 $_{\small{58.7}}$  &
            52.8 $_{\small{55.3}}$
            \\
    \textit{en-tr} &  58.7 $_{\small{52.0}}$ & 
                40.0  $_{\small{60.3}}$ &
                64.2  $_{\small{59.4}}$ &
                51.6  $_{\small{52.7}}$ &
                50.4  $_{\small{55.3}}$
                 \\
    \textit{en-el} &  62.0 $_{\small{56.7}}$& 
                49.8   $_{\small{62.3}}$&
                62.8   $_{\small{59.4}}$&
                50.8   $_{\small{55.0}}$&
                53.1  $_{\small{55.3}}$
                 \\
    \textit{en-es} &  71.1 $_{\small{68.9}}$& 
                41.7   $_{\small{72.9}}$&
                63.6   $_{\small{59.4}}$&
                56.3   $_{\small{58.7}}$&
                47.2  $_{\small{55.3}}$
                \\
    \textit{en-ar} &  46.5 $_{\small{56.5}}$& 
                68.5   $_{\small{64.1}}$&
                47.0   $_{\small{59.4}}$&
                57.5   $_{\small{55.0}}$&
                56.6  $_{\small{55.3}}$
                \\
    \midrule
        \textit{de-en} & 78.3 $_{\small{71.3}}$  & 
            46.3 $_{\small{64.1}}$  & 
            62.0 $_{\small{58.7}}$  & 
            47.0 $_{\small{59.4}}$  &
            56.6  $_{\small{55.3}}$
            \\
    \textit{de-tr} &  56.5 $_{\small{52.3}}$ & 
                48.7  $_{\small{53.7}}$ &
                62.7  $_{\small{58.7}}$ &
                55.8  $_{\small{52.7}}$ &
                54.3  $_{\small{55.3}}$
                 \\
    \textit{de-el} &  62.6 $_{\small{57.3}}$& 
                55.6   $_{\small{57.2}}$&
                62.5   $_{\small{58.7}}$&
                56.1   $_{\small{55.0}}$&
                55.8  $_{\small{55.3}}$
                 \\
    \textit{de-es} &  69.0 $_{\small{65.3}}$& 
                41.9   $_{\small{61.8}}$&
                61.8   $_{\small{58.7}}$&
                56.6   $_{\small{58.7}}$&
                50.0  $_{\small{55.3}}$
                \\
    \textit{de-ar} &  58.9 $_{\small{57.1}}$& 
                62.5   $_{\small{57.5}}$&
                57.3   $_{\small{58.7}}$&
                57.1   $_{\small{55.0}}$&
                56.9  $_{\small{55.3}}$
                \\
    \midrule
        \textit{tr-en} & 76.1 $_{\small{60.3}}$  & 
            38.7 $_{\small{52.0}}$  & 
            61.1 $_{\small{52.7}}$  & 
            51.9 $_{\small{59.4}}$  &
            56.9  $_{\small{55.3}}$
            \\
    \textit{tr-de} &  72.3 $_{\small{53.7}}$ & 
                37.3  $_{\small{52.3}}$ &
                62.0  $_{\small{52.7}}$ &
                45.4  $_{\small{58.7}}$ &
                57.4  $_{\small{55.3}}$
                 \\
    \textit{tr-el} &  63.4 $_{\small{52.9}}$& 
                41.7   $_{\small{50.8}}$&
                61.7   $_{\small{52.7}}$&
                55.2   $_{\small{55.0}}$&
                56.5  $_{\small{55.3}}$
                 \\
    \textit{tr-es} &  64.4 $_{\small{55.1}}$& 
                39.7   $_{\small{51.4}}$&
                59.9   $_{\small{52.7}}$&
                57.6   $_{\small{58.7}}$&
                53.5  $_{\small{55.3}}$
                \\
    \textit{tr-ar} &  54.8 $_{\small{52.5}}$& 
                53.1   $_{\small{50.5}}$&
                53.8   $_{\small{52.7}}$&
                55.9   $_{\small{55.0}}$&
                54.9  $_{\small{55.3}}$
                \\
    \midrule
        \textit{el-en} & 76.0 $_{\small{62.3}}$  & 
            40.8 $_{\small{56.7}}$  & 
            60.1 $_{\small{55.0}}$  & 
            46.2 $_{\small{59.4}}$  &
            54.9  $_{\small{55.3}}$
            \\
    \textit{el-de} &  72.8 $_{\small{57.2}}$ & 
                39.8  $_{\small{57.3}}$ &
                61.9  $_{\small{55.0}}$ &
                45.5  $_{\small{58.7}}$ &
                56.3  $_{\small{55.3}}$
                 \\
    \textit{el-tr} &  69.6 $_{\small{50.8}}$& 
                40.1   $_{\small{52.9}}$&
                59.7   $_{\small{55.0}}$&
                40.9   $_{\small{52.7}}$&
                54.3  $_{\small{55.3}}$
                 \\
    \textit{el-es} &  67.1 $_{\small{60.9}}$& 
                43.6   $_{\small{58.3}}$&
                60.4   $_{\small{55.0}}$&
                57.4   $_{\small{58.7}}$&
                54.1  $_{\small{55.3}}$
                \\
    \textit{el-ar} &  57.1 $_{\small{54.6}}$& 
                56.4   $_{\small{54.5}}$&
                57.6   $_{\small{55.0}}$&
                57.3   $_{\small{55.0}}$&
                56.0  $_{\small{55.3}}$
                \\
    \midrule
        \textit{es-en} & 79.3 $_{\small{72.9}}$  & 
            48.4 $_{\small{68.9}}$  & 
            61.4 $_{\small{58.7}}$  & 
            46.3 $_{\small{59.4}}$  &
            56.0  $_{\small{55.3}}$
            \\
    \textit{es-de} &  76.4 $_{\small{61.8}}$ & 
                43.1  $_{\small{65.3}}$ &
                63.2  $_{\small{58.7}}$ &
                43.8  $_{\small{58.7}}$ &
                56.4  $_{\small{55.3}}$
                 \\
    \textit{es-tr} &  71.9 $_{\small{51.4}}$& 
                39.1   $_{\small{55.1}}$&
                62.4   $_{\small{58.7}}$&
                40.6   $_{\small{52.7}}$&
                55.4  $_{\small{55.3}}$
                 \\
    \textit{es-el} &  74.7 $_{\small{58.3}}$& 
                38.1   $_{\small{60.9}}$&
                62.8   $_{\small{58.7}}$&
                38.1   $_{\small{55.0}}$&
                55.3  $_{\small{55.3}}$
                \\
    \textit{es-ar} &  42.4 $_{\small{57.3}}$& 
                70.1   $_{\small{61.0}}$&
                54.6   $_{\small{58.7}}$&
                59.6   $_{\small{55.0}}$&
                54.9  $_{\small{55.3}}$
                \\
    \midrule
        \textit{ar-en} & 75.9 $_{\small{64.1}}$  & 
            39.2 $_{\small{56.5}}$  & 
            59.6 $_{\small{55.0}}$  & 
            41.6 $_{\small{59.4}}$  &
            54.9  $_{\small{55.3}}$
            \\
    \textit{ar-de} &  72.1 $_{\small{57.5}}$ & 
                38.2  $_{\small{57.1}}$ &
                61.1  $_{\small{55.0}}$ &
                41.5  $_{\small{58.7}}$ &
                55.8  $_{\small{55.3}}$
                 \\
    \textit{ar-tr} &  69.3 $_{\small{50.5}}$& 
                38.0   $_{\small{52.5}}$&
                58.9   $_{\small{55.0}}$&
                37.7   $_{\small{52.7}}$&
                54.2  $_{\small{55.3}}$
                 \\
    \textit{ar-el} &  71.2 $_{\small{54.5}}$& 
                36.2   $_{\small{54.6}}$&
                61.0   $_{\small{55.0}}$&
                37.9   $_{\small{55.0}}$&
                55.5  $_{\small{55.3}}$
                \\
    \textit{ar-es} &  73.6 $_{\small{61.0}}$& 
                39.8   $_{\small{57.3}}$&
                60.8   $_{\small{55.0}}$&
                43.3   $_{\small{58.7}}$&
                54.6  $_{\small{55.3}}$
                \\
        
\bottomrule
    
    \end{tabular}}

    \caption{mBERT's performance on the XNLI test set. The columns show the fine-tuning language pairs, and the rows show the evaluation pairs in which $l1$ and $l2$ represent the premise and hypothesis's languages, respectively, as given in the corresponding columns. The rows including $*$ show the average performance over all languages. The smaller numbers present the baseline performance (fine-tuning on en-en) for the corresponding evaluation pairs.}
    \label{tab:xnli-mix-mbert}
\end{table}

\begin{table*}[ht!]
    \centering   
    \setlength{\tabcolsep}{6.5pt}
    \scalebox{0.8}{
    \begin{tabular}{r | c c c c c c c c | c c c c c c c |c}

    \toprule
              & en & de & fr & ru & es & zh & vi & ar & tr & bg & el & ur & hi & th & sw & avg \\
    \midrule          
    \multicolumn{17}{c}{\textbf{Entailment}}
    \\
    \it within & 81.6  & 
            66.9   & 
            70.1   & 
            65.5   & 
            70.5   &
            59.1   & 
            62.7   & 
            60.4   & 
            61.9   & 
            69.3   & 
            65.9   & 
            47.0   & 
            53.9   & 
            55.7   & 
            60.1   &
            63.4\\
    \it across &  54.5 & 
                49.4   &
                53.0   &
                50.9   &
                53.6   &
                42.5   &
                43.9   &
                41.3   &
                42.5   &
                52.4   &
                47.5   &
                36.5   &
                42.0   &
                41.9   &
                27.1   &
                45.3 \\

    \multicolumn{17}{c}{\textbf{NotEntailment}}
    \\
    \it within & 86.5  & 
            81.8   & 
            82.2   & 
            80.9   & 
            83.5   &
            80.8   & 
            80.6   &
            77.2   & 
            78.7   & 
            82.7   & 
            80.8   & 
            75.1   & 
            77.1   & 
            80.0   & 
            67.9   & 
            79.7\\
    
    \it across &  80.6 & 
                76.5   &
                77.1   &
                76.8   &
                77.2   &
                75.6   &
                76.1   &
                74.1   &
                73.4   &
                76.5   &
                76.3  &
                72.1   &
                74.7   &
                75.9   &
                67.5   &
                75.4\\
    \bottomrule

    \end{tabular}}

    \caption{Performance of XLM-r setting per label on the NLI task. Most of the performance drop in the \textit{across} setting occurs for the \textit{entailment} class.}
    \label{tab:xnli-perlabel-xlmr}
\end{table*}

\begin{table*}[ht!]
    \centering   
    \setlength{\tabcolsep}{6.5pt}
    \scalebox{0.8}{
    \begin{tabular}{r | c c c c c c c c | c c c c c c c |c}

    \toprule
              & en & de & fr & ru & es & zh & vi & ar & tr & bg & el & ur & hi & th & sw & avg \\
    \midrule          
    \multicolumn{17}{c}{\textbf{Entailment}}
    \\
    \it within & 82.4  & 
            68.0   & 
            70.8   & 
            66.1   & 
            72.0   &
            61.7   & 
            63.5   & 
            60.5   & 
            62.0   & 
            68.9   & 
            66.4   & 
            45.9   & 
            55.2   & 
            56.5   & 
            53.0   &
            63.5\\
    \it across &  64.8 & 
                54.6   &
                56.4   &
                54.8   &
                57.0   &
                49.2   &
                49.5   &
                48.2   &
                49.3   &
                55.8   &
                51.8   &
                39.4   &
                48.4   &
                48.7   &
                34.4   &
                50.8 \\

    \multicolumn{17}{c}{\textbf{NotEntailment}}
    \\
    \it within & 87.7  & 
            83.1   & 
            83.6   & 
            82.3   & 
            84.0   &
            82.7   & 
            82.2   &
            79.7   & 
            80.7   & 
            83.2   & 
            82.0   & 
            76.1   & 
            78.9   & 
            81.3   & 
            72.5   & 
            81.3\\
    
    \it across &  83.2 & 
                80.7   &
                81.1   &
                80.4   &
                81.3   &
                80.4   &
                80.1   &
                79.2   &
                78.9   &
                80.8   &
                80.3  &
                77.3   &
                78.9   &
                80.2   &
                75.2   &
                79.9\\
    \bottomrule

    \end{tabular}}

    \caption{The performance of I{\small{NFO}}XLM setting per label on the NLI task. Most of the performance drop in the \textit{across} setting occurs for the \textit{entailment} class.}
    \label{tab:xnli-perlabel-info}
\end{table*}

\begin{table}[ht!]
    \centering   
    \setlength{\tabcolsep}{6.5pt}
    \scalebox{0.7}{
    \begin{tabular}{r | c c c c c c c |c }

    \toprule
              & en & de & fr & es & zh & ko & ja & avg \\
    \midrule 
    \multicolumn{9}{c}{\textbf{Paraphrase}}
    \\

    \textit{within}
    & 95.8
    & 87.7
    & 89.6
    & 88.8
    & 80.9
    & 63.5
    & 71.8
    & 82.6 \\
    \textit{across}
    & 58.4
    & 47.3
    & 47.8
    & 47.9
    & 26.5 
    & 32.3 
    & 25.2
    & 40.8 \\
    \midrule 
    \multicolumn{9}{c}{\textbf{NonParaphrase}}
    \\
    \textit{within}
    & 93.2
    & 87.8
    & 89.4
    & 89.4
    & 82.7
    & 86.7
    & 81.7
    & 87.3\\
    \textit{across}
    & 91.1
    & 92.2
    & 92.0
    & 91.9
    & 93.6
    & 91.8
    & 93.2
    & 92.3\\
    
\bottomrule
\end{tabular}}

    \caption{Performance of XLM-r for the \textit{within} and \textit{across} setting per label on the PAWS-X test set. Most of the performance drop of the \textit{across} setup setting originates from the drop in the \textit{Paraphrase} class.}
    \label{tab:paws-xlmr-performance}
\end{table}

\begin{table}[ht!]
    \centering   
    \setlength{\tabcolsep}{6.5pt}
    \scalebox{0.7}{
    \begin{tabular}{r | c c c c c c c |c }

    \toprule
              & en & de & fr & es & zh & ko & ja & avg \\
    \midrule 
    \multicolumn{9}{c}{\textbf{Paraphrase}}
    \\

    \textit{within}
    & 96.0
    & 92.4
    & 92.7
    & 91.2
    & 86.8
    & 74.8
    & 81.3
    & 87.9 \\
    \textit{across}
    & 84.5
    & 77.4
    & 77.4
    & 76.8
    & 70.3 
    & 65.0 
    & 67.8
    & 74.2 \\
    \midrule 
    \multicolumn{9}{c}{\textbf{NonParaphrase}}
    \\
    \textit{within}
    & 92.9
    & 85.1
    & 88.3
    & 90.0
    & 79.8
    & 82.2
    & 77.3
    & 85.1\\
    \textit{across}
    & 85.5
    & 81.6
    & 82.8
    & 83.2
    & 78.8
    & 78.4
    & 77.5
    & 81.1\\
    
\bottomrule
\end{tabular}}

    \caption{Performance of I{\small{NFO}}XLM for the \textit{within} and \textit{across} setting per label on the PAWS-X test set. Most of the performance drop of the \textit{across} setup setting originates from the drop in the \textit{Paraphrase} class.}
    \label{tab:paws-info-performance}
\end{table}

\begin{table*}[ht!]
    \centering   
    \setlength{\tabcolsep}{6.5pt}
    \scalebox{0.7}{
    \begin{tabular}{r | c c c c c c c c c c c c c c c c c|c}

    \toprule
              & en & de & fr & ru & es & zh & vi & ar & tr & bg & el & ur & hi & th & sw & ko & ja & avg \\
    \midrule          
    \multicolumn{19}{c}{\textbf{XNLI}}
    \\
    \cmidrule{2-18}
    \it within & 81.3  & 
            74.1   & 
            75.8   & 
            72.9   & 
            76.0   &
            71.0   & 
            71.0   &
            69.5   & 
            71.3   & 
            75.2   & 
            72.9   & 
            63.8   & 
            66.7   & 
            68.8   & 
            62.6   & 
            ~-~    &
            ~-~    &

            71.3 \\
    \it across &  68.1 & 
                63.7   &
                65.2   &
                65.1   &
                65.2   &
                61.6   &
                60.4   &
                59.9   &
                60.1   &
                64.9   &
                63.4   &
                57.1   &
                60.2   &
                61.8   &
                51.7   &
                ~-~    & 
                ~-~    &
                61.9 \\

    \cmidrule{2-18}
    \multicolumn{19}{c}{\textbf{PI}}
    \\
    \cmidrule{2-18}
    \it within & 54.6  & 
            55.2   & 
            54.8   & 
            ~-~    & 
            54.6   &
            55.3   & 
            ~-~    & 
            ~-~    & 
            ~-~    & 
            ~-~    & 
            ~-~    & 
            ~-~    & 
            ~-~    & 
            ~-~    & 
            ~-~    &
            55.2   &
            55.8   &
            55.1\\
    \it across &  55.0 & 
                55.2   &
                55.1   &
                ~-~    &
                55.0   &
                55.2   &
                ~-~    &
                ~-~    &
                ~-~    &
                ~-~    &
                ~-~    &
                ~-~    &
                ~-~    &
                ~-~    &
                ~-~    &
                55.2   &
                55.5   &
                55.2\\
    \cmidrule{2-18}
        \multicolumn{19}{c}{\textbf{QA}}
    \\
    \cmidrule{2-18}
    \it within & 82.8  & 
            73.6   & 
            ~-~    & 
            72.8   & 
            76.1   &
            62.7   & 
            72.7   & 
            63.9   & 
            66.1   & 
            ~-~    & 
            71.4   & 
            ~-~    & 
            66.6   & 
            66.7   & 
            ~-~    &
            ~-~    &
            ~-~    &
            70.5 \\
    \it across &  56.6 & 
                44.1   &
                ~-~    &
                43.5   &
                42.9   &
                30.2   &
                34.7   &
                27.8   &
                34.7   &
                ~-~    &
                37.5   &
                ~-~    &
                34.7   &
                34.1   &
                ~-~    &
                ~-~        &
                ~-~        &
                38.3\\
    \bottomrule

    \end{tabular}}

    \caption{Performance of XLM-r fine-tuned on the control tasks. We report the accuracy score for the NLI and PI tasks and the F1 score for QA. Although the fine-tuning data does not train the model with any task-related knowledge, the drop in the performance is negligible.}
    \label{tab:shuffle-performance}
\end{table*}

\begin{figure*}
    \centering
    \scalebox{0.9}{
    \includegraphics[height = 6cm, width=14.5cm]{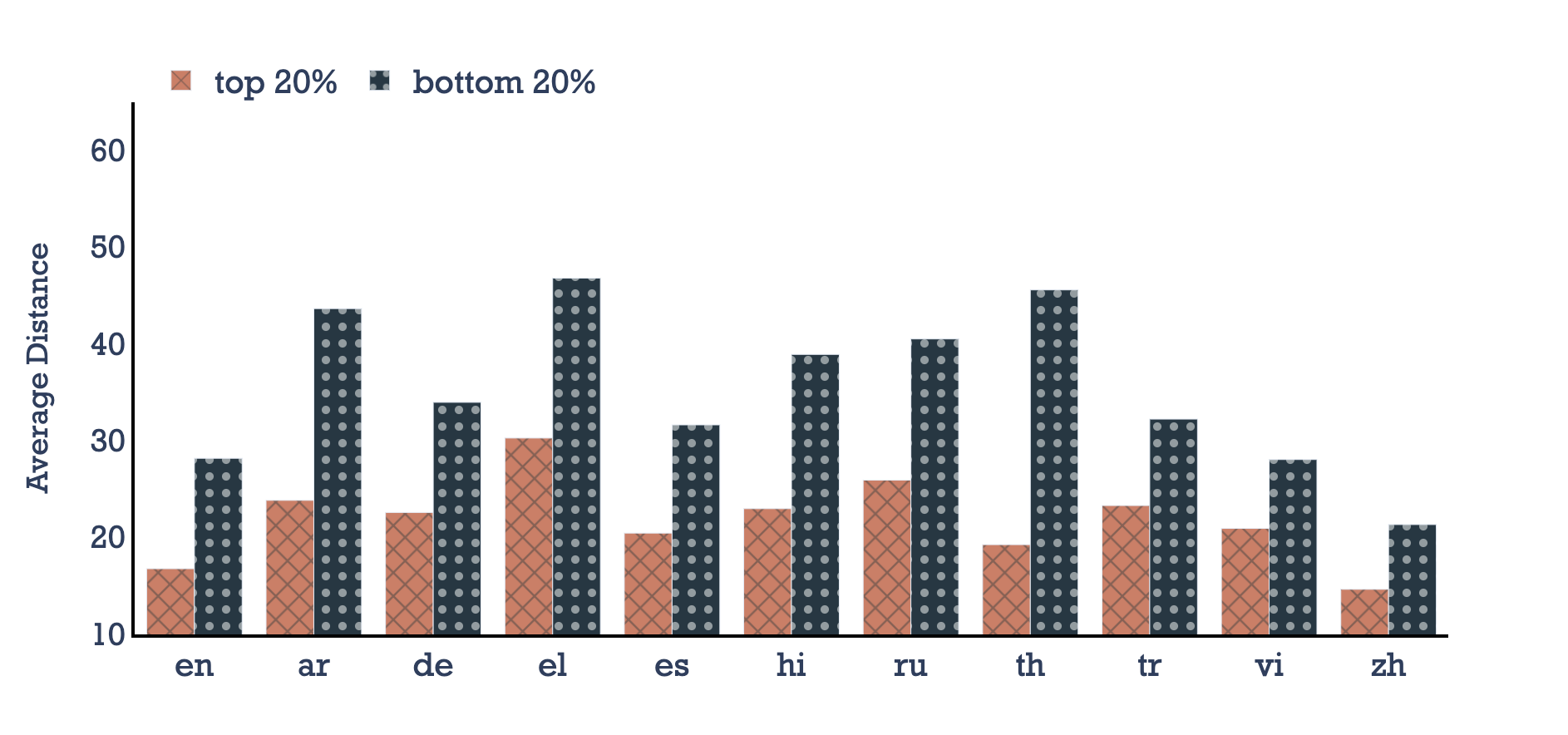}}
    \caption{The average distance of the questions' words occurred in the context to the center of the answer span in the top $20\%$ easiest and hardest instances for XLM-r fine-tuned on SQuAD based on the test set of every language. }
    \label{fig:qa-xlmr-distance}
\end{figure*}

\begin{figure*}
    \centering
    \scalebox{0.9}{
    \includegraphics[height = 6cm, width=14.5cm]{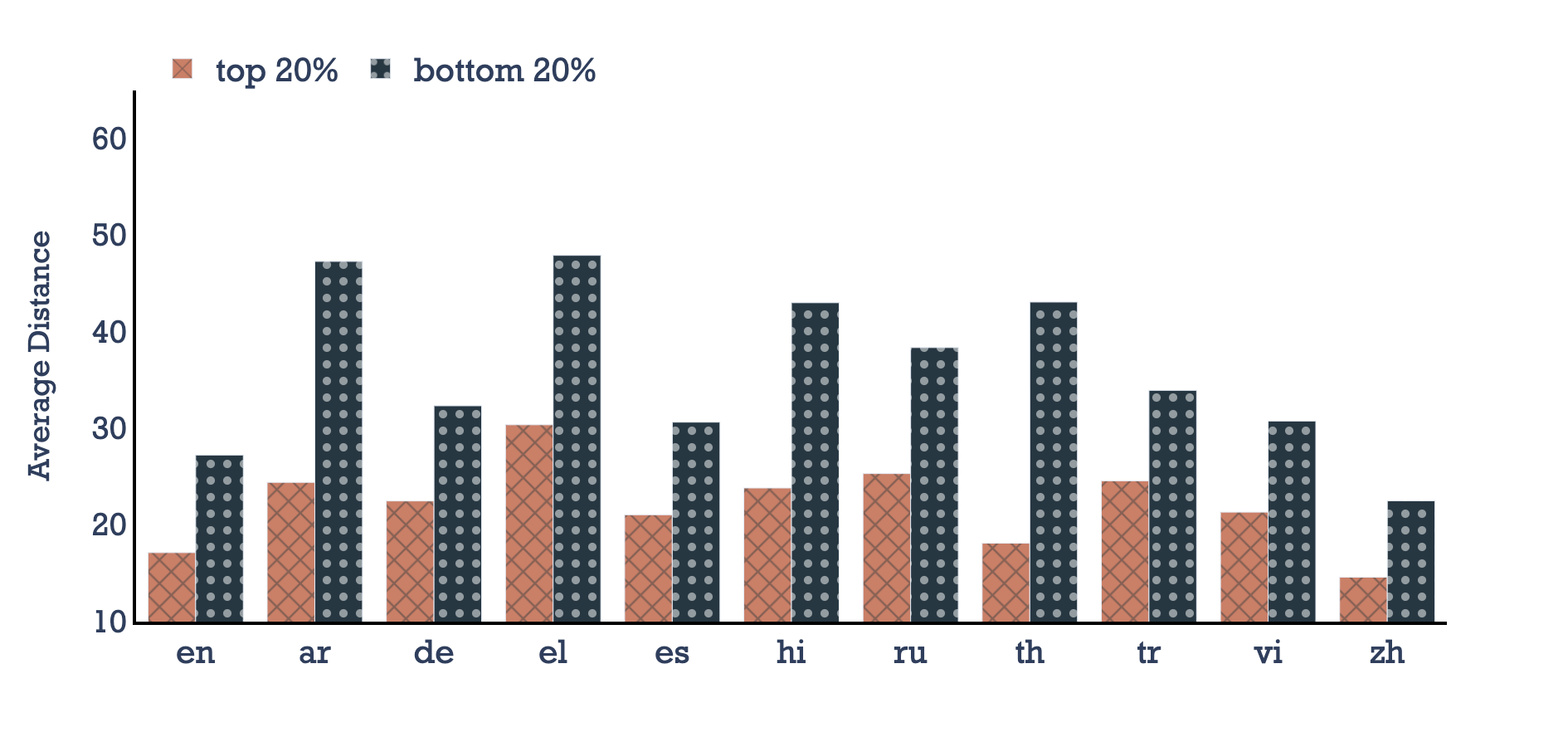}}
    \caption{The average distance of the questions' words occurred in the context to the center of the answer span in the top $20\%$ easiest and hardest instances for I{\small{NFO}}XLM fine-tuned on SQuAD based on the test set of every language. }
    \label{fig:qa-info-distance}
\end{figure*}

\end{document}